\documentclass[runningheads,a4paper]{llncs}
\usepackage{amssymb}
\usepackage{wrapfig,lipsum,booktabs}
\usepackage{graphicx}
\usepackage{algorithm}
\usepackage{algorithmicx}
\usepackage{algpseudocode}
\usepackage{amsmath}
\usepackage{multicol}
\usepackage{multirow}
\usepackage{epstopdf}
\usepackage{array}
\newcolumntype{C}[1]{>{\centering\let\newline\\\arraybackslash\hspace{0pt}}m{#1}}
\floatname{algorithm}{Algorithm}
\newcommand{\keywords}[1]{\par\addvspace\baselineskip
\noindent\keywordname\enspace\ignorespaces#1}

\begin{document}
\mainmatter  

\title{Large-scale Feature Selection of Risk Genetic Factors for Alzheimer's Disease via Distributed Group Lasso Regression}
\titlerunning{Large-scale Feature Selection of Risk Genetic Factors}

%
%
\author{Qingyang Li$^1$, Dajiang Zhu$^2$, Jie Zhang$^1$, Derrek Paul Hibar$^2$, Neda Jahanshad$^2$, Yalin Wang$^1$, Jieping Ye$^3$, Paul M. Thompson$^2$, Jie Wang$^3$}
\authorrunning{Q. Li et al.}

\institute{
$^1$School of Computing, Informatics, and Decision Systems Engineering, Arizona State Univ.,Tempe, AZ; $^2$Imaging Genetics Center, Institute for Neuroimaging and Informatics, Univ.of Southern California, Marina del Rey, CA; $^3$Dept. of Computational Medicine and Bioinformatics, Univ. of Michigan, Ann Arbor, MI}

\toctitle{Large-scale Feature Selection of Risk Genetic Factors}
\tocauthor{Authors' Instructions}
\maketitle

\begin{abstract}
Genome-wide association studies (GWAS) have achieved great success in the genetic study of Alzheimer's disease (AD). Collaborative imaging genetics studies across different research institutions show the effectiveness of detecting genetic risk factors. However, the high dimensionality of GWAS data poses significant challenges in detecting risk SNPs for AD. Selecting relevant features is crucial in predicting the response variable. In this study, we propose a novel Distributed Feature Selection Framework (DFSF) to conduct the large-scale imaging genetics studies across multiple institutions. To speed up the learning process, we propose a family of distributed group Lasso screening rules to identify irrelevant features and remove them from the optimization. Then we select the relevant group features by performing the group Lasso feature selection process in a sequence of parameters. Finally, we employ the stability selection to rank the top risk SNPs that might help detect the early stage of AD. To the best of our knowledge, this is the first distributed feature selection model integrated with group Lasso feature selection as well as detecting the risk genetic factors across multiple research institutions system. Empirical studies are conducted on 809 subjects with 5.9 million SNPs which are distributed across several individual institutions, demonstrating the efficiency and effectiveness of the proposed method.

\end{abstract}

\section{Introduction}

Alzheimer’s disease (AD) is known as the most common type of dementia. Genome-Wide Association Studies (GWAS) \cite{harold2009genome} achieved great success in finding single nucleotide polymorphisms (SNPs) associated with AD. Some large-scale collaborative network such as ENIGMA \cite{thompson2014enigma} Consortium consists of 185 research institutions around the world, analyzing genomic data from over 33,000 subjects, from 35 countries. However, processing and integrating genetic data across different institutions is challenging. The first issue is the data privacy since each participating institution wishes to collaborate with others without revealing its own data set. The second issue is how to conduct the learning process across different institutions. Local Query Model (LQM) \cite{li2016large,zhudajiang2017large} is proposed to perform the distributed Lasso regression for large-scale collaborative imaging genetics studies across different institutions while preserving the data privacy for each of them. However, in some imaging genetics studies  \cite{harold2009genome}, we are more interested in finding important explanatory factors in predicting responses, where each explanatory factor is represented by a group of features since lots of AD genes are continuous or relative with some other features, not individual features. In such cases, the selection of important features corresponds to the selection of groups of features. As an extension of Lasso, group Lasso \cite{yuan2006model} has been proposed for feature selection in a group level  and quite a few efficient algorithms \cite{qin2013efficient,boyd2011distributed} have been proposed for efficient optimization. However, integrating group Lasso with imaging genetics studies across multiple institutions has not been studied well. 


In this study, we propose a novel Distributed Feature Selection Framework (DFSF) to conduct the large-scale imaging genetics studies analysis across multiple research institutions. Our framework has three components. In the first stage, we proposed a family of distributed group lasso screening rules (DSR and DDPP\_GL) to identify inactive features and remove them from the optimization. The second stage is to perform the group lasso feature selection process in a distributed manner, selecting the top relevant group features for all the institutions. Finally, each institution obtains the learnt model and perform the stability selection to rank the top risk genes for AD. The experiment is conducted on the Alzheimer's Disease Neuroimaging Initiative (ADNI) GWAS data set, including approximately 809 subjects with 5.9 million loci. Empirical studies demonstrate that proposed method the proposed method achieved a 35-fold speedup compared to state-of-the-art distributed solvers like ADMM. Stability selection results show that the proposed DFSF detects \textit{APOE}, \textit{GRM8}, \textit{GPC6} and \textit{LOC100506272} as top risk SNPs associated with AD, demonstrating a superior result compared to Lasso regression methods \cite{li2016large}. The proposed method offers a powerful feature selection tool to study AD and its early symptom.

\section{Problem Statement}

\subsection{Problem Formulation}

Group Lasso \cite{yuan2006model} is a highly efficient feature selection and regression technique used in the model construction. Group Lasso takes the form of the equation:

\begin{equation}
 \min\limits_{x\in  \mathbb{R}^N } {F(x)=\frac{1}{2}||y-\sum_{g=1}^{G}[A]_g [x]_g||_2^ 2+ \lambda \sum_{g=1}^{G} w_g ||[x]_g||_2,   }
\label{eq:1}
\end{equation}
where $A$ represents the feature matrix where $A\in \mathbb{R}^{N\times P}$ and y denotes the $N$ dimensional response vector. $\lambda$ is a positive regularization parameter.
Different from Lasso regression \cite{tibshirani1996regression}, group Lasso partitions the original feature matrix $A$ into $G$ non-overlapping groups $[A]_1, [A]_2,......,[A]_G$ and $w_g$ denotes the weight for the $g$-th group. After solving the group Lasso problem, we get the corresponding $G$ solution vector $[x]_1, [x]_2,......,[x]_G$ and the dimension of $[x]_g$ is the same as the feature space in $[A]_g$.

\subsection{ADNI GWAS data}

The ADNI GWAS dataset contains genotype information of 809 ADNI participants. To store statistically relevant SNPs called using Illumina’s CASAVA SNP Caller, the ADNI WGS SNP data is stored in variant call format (VCF) for storing gene sequence variations. SNPs at approximately 5.9 million specific loci are recorded for each participant. We encode SNPs using the coding scheme in \cite{sasieni1997genotypes} and apply Minor Allele Frequency (MAF) $<0.05$ and Genotype Quality (GQ) $<45$ as two quality control criteria to filter high quality SNPs features. We follow the same SNP genotype coding and quality control scheme in \cite{li2016large}. 

We have $m$ institutions to conduct the collaborative learning. The $i$th institution maintains its own data set $(A_i,y_i)$ where $A_i\in \mathbb{R}^{n_i\times P}$, $n_i$ is the sample number, $P$ is the feature number and $y_i\in \mathbb{R}^{n_i}$ is the response and $N=\sum_i^m n_i$. We assume $P$ is the same across $m$ institutions. We aim at conducting the feature selection process of group lasso on the distributed datasets $(A_i,y_i)$, $i=1,2,...,m$.

\begin{figure}[!t]
\centering
\includegraphics[height=5cm]{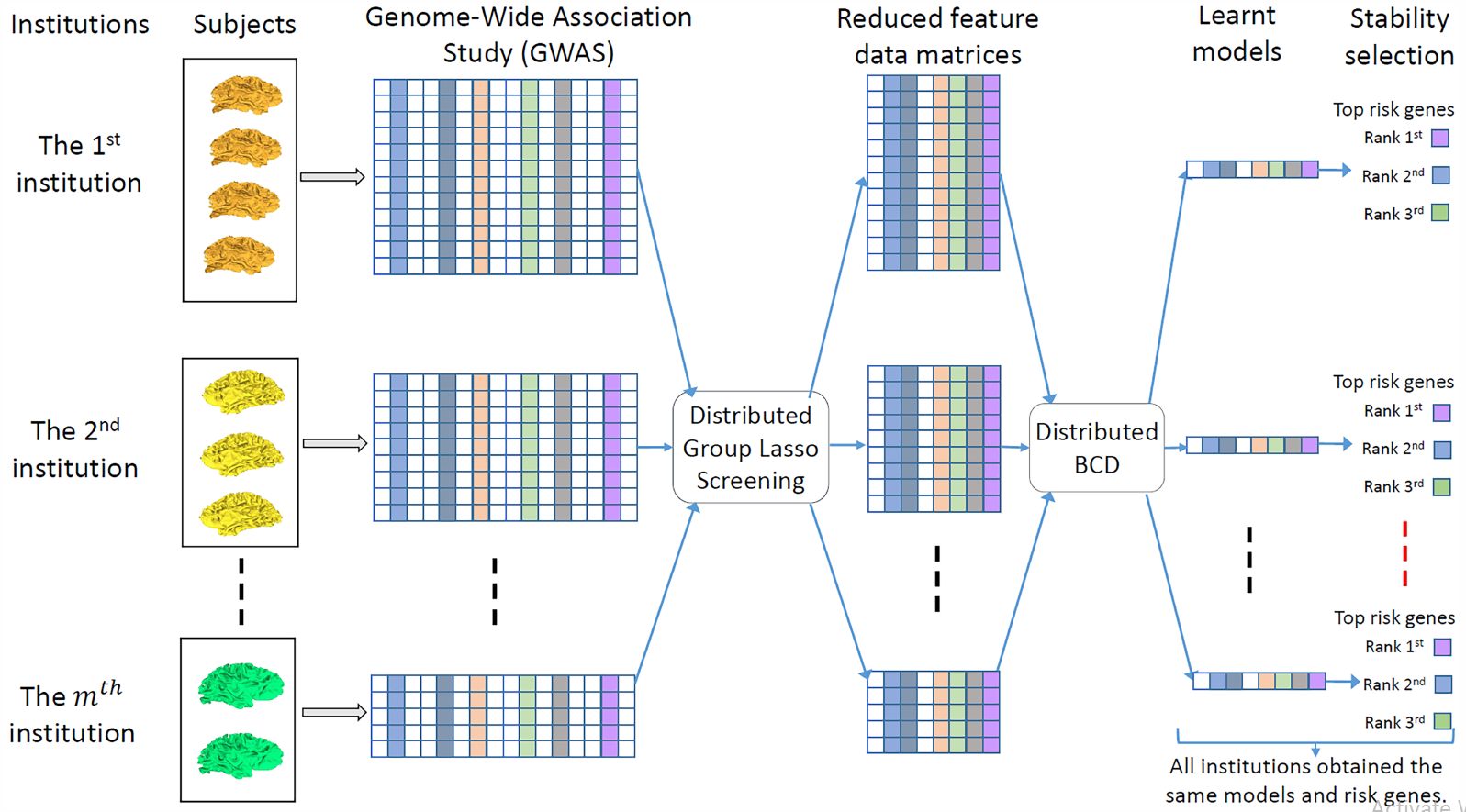}
\caption{Illustration of our DFSF framework. Each participating institution maintains its own dataset which are a few subjects of GWAS dataset. Firstly, we perform the distributed group Lasso screening rules to pre-identifying the inactive features and remove them from the optimization. Then we conduct the learning process of group Lasso by proposed distributed solver DBCD to select the top relevant features. Finally, each institution obtains the same selected features and performs stability selection to rank the top SNPs that may collectively affect AD. 
}
\label{fig:1}
\end{figure}

\section{Proposed Framework}

In this section, we present the streamline of proposed DFSF framework. The DFSF framework is composed of three main procedures:
\begin{enumerate}
\item Identify the inactive features by the distributed group Lasso screening rules and remove inactive features from optimization.
\item Solve the group Lasso problem on the reduced feature matrix along a sequence of parameter values and select the most relevant features for each participating institution.
\item Perform the stability selection to rank SNPs that may collectively affect AD.
\end{enumerate}

\subsection{Screening Rules for Group Lasso}

Strong rule \cite{strongrules} is an efficient screening method for fitting lasso-like problems by pre-identifying the features which have zero coefficients in the solution and removing these features from optimization, significantly cutting down on the computation required for optimization.

For the group lasso problem \cite{yuan2006model}, the $g$th group of $x$---$[x]_g$--- will be discarded by strong rules if the following rule holds:
\begin{equation}
||[A]_g^T y||_2\leq w_g(2\lambda-\lambda_{\max})
\label{eq:2}
\end{equation}
The calculation of $\lambda_{\max}$ follows $\lambda_{\max}=\max_{g}\frac{||[A]_g^T y||_2}{w_g}$. $[x]_g$ could be discarded in the optimization without sacrificing the accuracy since all the elements of $[x]_g$ are zero in the optimal solution vector. 

Let $J$ denote the index set of groups in the feature space and $J=\{1, 2,......,G \}$. Suppose that there are $\widetilde{G}$ remaining groups after employing screening rules, we use $\widetilde{J}$ to represent the index set of remaining groups and $\widetilde{J}=\{1, 2,......,\widetilde{G} \}$. As a result, the optimization of group lasso problem (\ref{eq:1}) can be reformulated as:
\begin{equation}
 \min\limits_{\widetilde{x}\in  \mathbb{R}^{\widetilde{N}} } {F(\widetilde{x})=\frac{1}{2}||y-\sum_{g=1}^{\widetilde{G}}[\widetilde{A}]_g [\widetilde{x}]_g||_2^ 2+ \lambda \sum_{g=1}^{\widetilde{G}} w_g ||[\widetilde{x}]_g||_2,   }
\label{eq:3}
\end{equation}
where $\widetilde{N}$ is the dimension of reduced feature space and $\widetilde{x}\in  \mathbb{R}^{\widetilde{N}}$.

\subsection{Distributed Screening Rules for Group Lasso}
As the data set are distributed among multiple research institutions, it is necessary to conduct a distributed learning process without compromising the data privacy for each institution. LQM \cite{li2016large,zhudajiang2017large} is proposed to optimize the lasso regression while preserving the data privacy for each participating institution. In this study, we aim at selecting the group features to detect the top risk genetic factors for the entire GWAS data set. Since each institution maintains its own data pair $(A_i, y_i)$, we develop a family of distributed group Lasso screening to identify and discard the inactive features in a distributed environment. We summarize the Distributed Strong Rules (DSR) as follows:

\begin{enumerate}
  \item For the $i$th institution, compute $Q_i$ by $Q_i=A_i^T y_i$.
  \item Update $Q = \sum_i^m Q_i$ by LQM, then send $Q$ back to all the institutions.
  \item In each institution, calculate $\lambda_{\max}$ by: $\lambda_{\max}=\max_g \frac{||[Q]_g||_2}{w_g} $ where $[Q]_g$ is the elements of $g$th group in $Q$ and it is similar as the definition of $[A]_g$.
   \item For each $g$th group in the problem (\ref{eq:1}), we will discard it and remove from the optimization when the following rule holds:  $||[Q]_g||_2\leq w_g(2\lambda-\lambda_{\max}).$
\end{enumerate}

In many real word applications, the optimal value of regularization parameter $\lambda$ is unknown. To tune the value of $\lambda$, commonly used methods such as cross validation needs to solve the Lasso problem along a sequence of parameter values $\lambda_0>\lambda_1>...>\lambda_{\kappa}$ ,which can be very
time-consuming. A sequential version of strong rules was proposed in EDPP \cite{wang2013lasso} by utilizing the information of optimal solutions in the previous parameter, achieving about 200x speedups for real-world applications. The implementation details of EDPP is available on the GitHub: http://dpc-screening.github.io/glasso.html. We omit the introduction of EPDD for brevity. We propose a distributed safe screening rules for group Lasso, known as the Distributed Dual Polytope Projection Group Lasso (DDPP\_GL), to quickly identify and discard inactive features along a sequence of parameters in a distributed manner. We summarize DDPP\_GL in algorithm~\ref{alg:1}.

\begin{algorithm}[t]
\caption{Distributed Dual Polytope Projection for Group Lasso}
\begin{algorithmic}[1]
   \Require A set of data pairs $\{(A_1, y_1), (A_2, y_2),..., (A_m, y_m)\}$ and $i$th institution holds the data pair $(A_i, y_i)$. A sequence of parameters: $\lambda_{\max} = \lambda_0 >...> \lambda_{\kappa}$.
   \Ensure The learnt models: $\{x^*(\lambda_0),x^*(\lambda_1),...,x^*(\lambda_{\kappa})\}$.
   \smallskip
   \State Let $R_i=A_i^Ty_i$, compute $R=\sum_{i=1}^{m} R_i$ by LQM.
   \State $\lambda_{max}=\max_g \frac{||[R]_g||_2}{w_g}$, $[R]_g$ represents all the elements in the $g$th group.
   \State $S_i = argmax_{[A_i]_g}  \frac{||R_g||_2}{w_g} $, compute $L=\sum_{i=1}^{m} S_i^T y_i$ by LQM.
   \State Let $\lambda_0\in(0,\lambda_{\max}]$ and $\lambda\in(0, \lambda_0]$.
  \smallskip
   \State \quad $\theta_i(\lambda)  = \left\{
        \begin{array}{l}
        \frac{y_i-\sum_{g=1}^{G}[A_i]_g [x^*(\lambda)]_g}{\lambda}, \text{ if } \lambda \in (0, \lambda_{max}).\\
        \frac{y_i}{\lambda_{\max}}, \quad \quad \quad \quad \quad \quad \quad \text{ if } \lambda=\lambda_{max}.
        \end{array}
        \right. $
   \smallskip
   \State \quad $v_1(\lambda_0)_i  = \left\{
        \begin{array}{l}
        \frac{y_i}{\lambda_0}-\theta_i(\lambda_0), \quad \text{if } \lambda \in (0, \lambda_{max}) ,\\
        S_i L,\quad \quad \quad \quad \text{ if } \lambda=\lambda_{max}
        \end{array}
        \right.$
   \smallskip
   \State \quad $v_2(\lambda, \lambda_0)_i=\frac{y_i}{\lambda}-\theta_i(\lambda_0)$
   \smallskip
   \State \quad $Q_i=||v_1(\lambda_0)_i||_2^2$, compute $Q=\sum_i^m Q_i$ by LQM.
   \smallskip
   \State \quad $v_2^{\perp}(\lambda, \lambda_0)_i=v_2(\lambda,\lambda_0)_i-\frac{<v_1(\lambda_0)_i, v_2(\lambda,\lambda_0)_i>}{Q} v_1(\lambda_0)_i$
   \smallskip
   \State Given a sequence of parameters $\lambda_{\max}=\lambda_0>...>\lambda_{\kappa}$, for any integer $k\in [1, \kappa]$, we make a pre-screen on each groups of $[x^*(\lambda_k)]_g$, if $[x^*(\lambda_{k-1})]_g$ is known.
   \smallskip
   \State \quad \textbf{for} $g=1$ \textbf{to} $G$ \textbf{do}
   \State \quad \quad $Q_i=[A_i]^T_g(\theta^*(\lambda_{k-1})_i+\frac{1}{2}v_2^{\perp}(\lambda_k, \lambda_{k-1})_i)||_2$
   \smallskip
   \State \quad \quad Compute $Q=\sum_i^m Q_i$ by LQM.
   \smallskip
   \State \quad \quad \textbf{if} $Q<1-\frac{1}{2}||v_2^{\perp}(\lambda_k, \lambda_{k-1})||_2||[A]_g||_2$ \textbf{then}
   \smallskip
   \State \quad \quad \quad We identify all the elements of $[x^*(\lambda_k)]_g$ to be zero.
    \State \quad  \textbf{end for}
\end{algorithmic}
\label{alg:1}
\end{algorithm}



\subsection{Distributed Block Coordinate Descent for Group Lasso}

After we apply DDPP\_GL to discard the inactive features, the feature space shrank from $P$ to $\widetilde{P}$ and there are remaining $\widetilde{G}$ groups. The problem of group Lasso (\ref{eq:1}) could be reduced as (\ref{eq:3}). We need to optimize (\ref{eq:3}) in a distributed manner. The block coordinate descent (BCD) \cite{qin2013efficient} is one of the most efficient solvers in the big data optimization. BCD optimize the problem by updating one or a few blocks of variables at a time, rather than updating all the block together. The order of update can be deterministic or stochastic. For the group lasso problem, we can randomly pick up a group of variables to optimize and keeps other groups fixed. Following this idea, we propose a Distributed
Block\begin{algorithm}
\caption{Distributed Block Coordinate Descent}
\begin{algorithmic}[1]
   \Require A set of data pairs $\{(\widetilde{A}_1, y_1), (\widetilde{A}_2, y_2),..., (\widetilde{A}_n, y_n)\}$ where $i$th institution holds the data pair $(\widetilde{A}_i, y_i)$ and $\lambda$
   \Ensure The learnt models: $\widetilde{x}$.
   \State \textbf{Initialize:} $ \widetilde{x}= \textbf{0} \in \mathbb{R}^{\widetilde{P}} $ and $R_i=-y_i$.
   \State \textbf{while} not converged \textbf{do}
   \State \quad  Randomly pick up $g$ from the index set $\{1,..., \widetilde{G}\}$.
   \State \quad  Compute the $g$th group's gradient: $\nabla F([\widetilde{x}]_g)_i=[\widetilde{A}_i]_g^T R_i.$
   \State \quad  Update $\nabla F([\widetilde{x}]_g)$ by LQM: $\nabla F([\widetilde{x}]_g) = \sum_i^m \nabla F([\widetilde{x}]_g)_i$.
   \smallskip
   \State \quad  Let $v=[\widetilde{x}]_g$ and $[\widetilde{x}]_g = [\widetilde{x}]_g - \frac{1}{L_g}\nabla F([\widetilde{x}]_g)$
    \smallskip
   \State \quad $[\widetilde{x}]_g  = \left\{
        \begin{array}{l}
        [\widetilde{x}]_g - \frac{\lambda w_g}{||[\widetilde{x}]_g||_2}  [\widetilde{x}]_g, \quad \text{if } ||[\widetilde{x}]_g||_2 > \frac{\lambda w_g}{L_g}.\\
        \textbf{0} \in \mathbb{R}^{\widetilde{N}}, \quad \quad \quad \quad \quad \text{if } ||[\widetilde{x}]_g||_2 \leq \frac{\lambda w_g}{L_g}.
        \end{array}
        \right.$
   \smallskip
   \State \quad  Let compute $\Delta [\widetilde{x}]_g$ by:  $\Delta [\widetilde{x}]_g = [\widetilde{x}]_g-v$.
   \smallskip
   \State \quad Update $R_i$ by: $R_i=R_i+\Delta [\widetilde{x}]_g [\widetilde{A}_i]_g^T $
   \State \textbf{end while}
\end{algorithmic}
\label{alg:2}
\end{algorithm}
Coordinate Descent (DBCD) to solve the group Lasso problem in algorithm~\ref{alg:2}.

In algorithm~\ref{alg:2}, we use a variable $R_i$ to store the result of $\widetilde{A}_i\widetilde{x}-y_i$. $R_i$ is initialized as $-y_i$ since $\widetilde{x}$ is initialized to be zero at the beginning. In DBCD, the update of gradient can be divided as three steps:
\begin{enumerate}
  \item Compute the gradient: $\nabla F([\widetilde{x}]_g)_i=[\widetilde{A}_i]_g^T R_i $ and get $\nabla F([\widetilde{x}]_g)$ by LQM. 
  \item Get $\Delta [\widetilde{x}]_g$ by the gradient information $\nabla F([\widetilde{x}]_g)$. 
  \item Update $R_i$:   $R_i=R_i+\Delta [\widetilde{x}]_g [\widetilde{A}_i]_g^T $
\end{enumerate}

The update of $[\widetilde{x}]_g$ follow the equations in $7$rd line of algorithm~\ref{alg:2}. We update $[\widetilde{x}]_g$ if $||[\widetilde{x}]_g||_2$ is larger than $\frac{\lambda w_g}{L_g}$, otherwise all the elements of $[\widetilde{x}]_g$ are set to be zero. $L_g$ denotes the Lipschitz constant in $g$th group. For the group Lasso problem, $L_g$ is set to be $||[A]_g||_2^2$. DBCD updates $R_i$ at the end of each iteration to make sure $R_i$ stores the correct information of $\widetilde{A}_i\widetilde{x}-y_i$ in each iteration. 

\subsection{Feature selection by Group Lasso}
Given a sequence of parameter values: $\lambda_0>...>\lambda_{\kappa}$, we can obtain a sequence of learnt models $\{x^*(\lambda_0),...,x^*(\lambda_{\kappa})\}$ by employing DDPP\_GL+DBCD. For each group $g$ in the feature space $G$, we count the frequency of nonzero entries in the learnt model and rank the frequency by descent to get the top relevant features. We summarize the top $K$ feature selection process as follows:
\begin{enumerate}
  \item For each group $g$ in the feature space $G$, $I_g = I_g+1$, If $[x^*(\lambda_k)]_g$ is not equal to zero where $k\in (0, \kappa)$ and $I \in \mathbb{R}^G$. 
  \item Rank $I$ by descent and select the top $K$ relevant features from $A_i$ to construct the feature matrix $\bar{A_i}$.
\end{enumerate}

After obtaining the relevant features, we perform the stability selection \cite{li2016large,meinshausen2010stability} to rank the top genetic factors that are associated with the disease AD.

\section{Experimental Results}
In this section, we conduct several experiments to evaluate the efficiency and effectiveness of our methods. The proposed framework is implemented across three institutions with thirty computation nodes on Apache Spark: http://spark.apach-e.org, a state-of-the-art distributed computing platform. We perform DDPP\_GL+ DBCD on a sequence of parameter values and employ stability selection with our methods to determine top risk SNPs related to AD.

\begin{figure}[t]
\centering
\includegraphics[height=6cm]{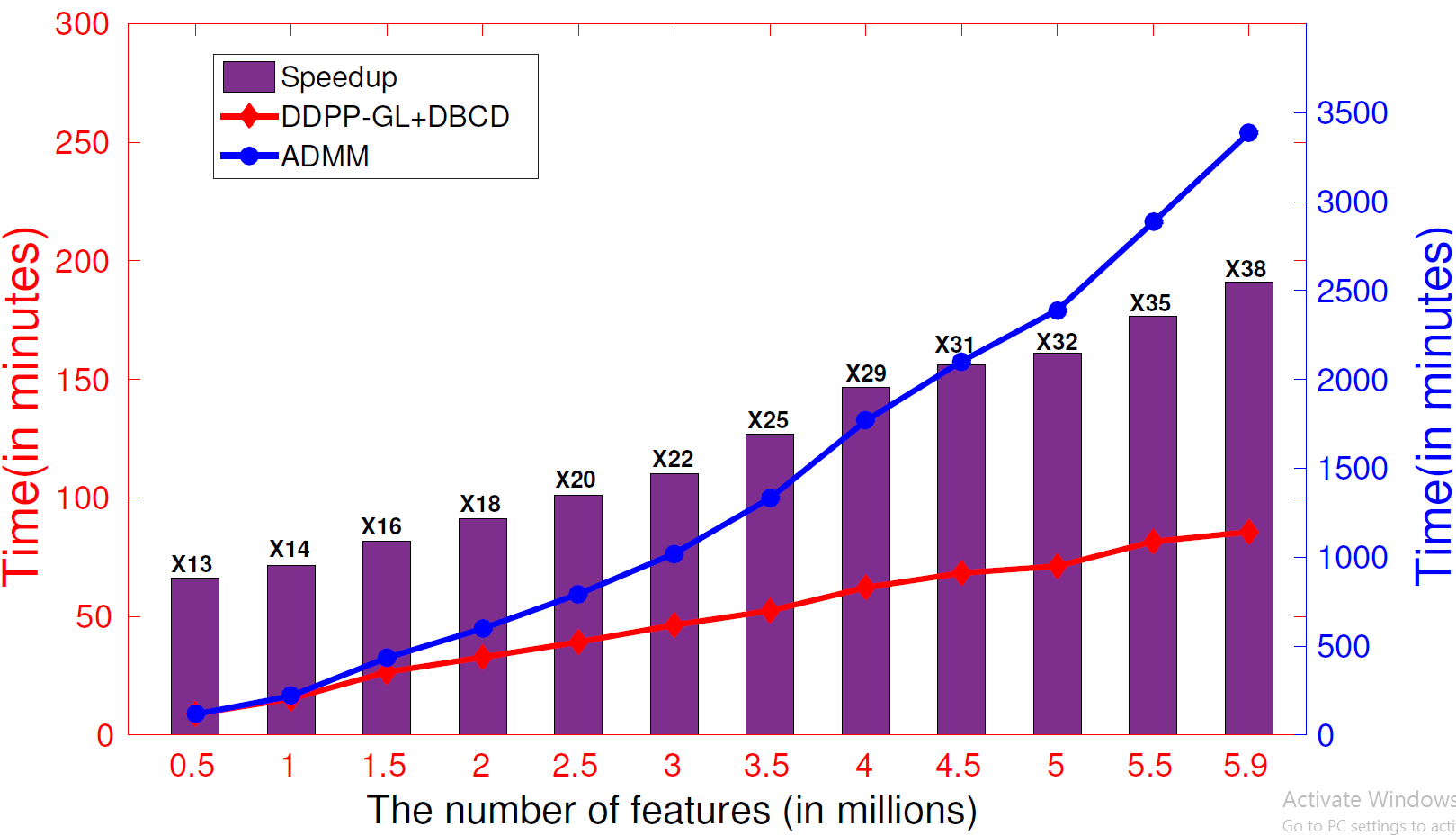}
\caption{Running time comparison of DDPP\_GL+DBCD with ADMM.}
\label{fig:2}
\end{figure}

\subsection{Performance Comparison}

In this experiment, we choose the volume of lateral ventricle as variables being predicted which  containing 717 subjects by removing subjects without labels. The volumes of brain regions were extracted from each subject's T1 MRI scan using Freesurfer: http://freesurfer.net. The distributed platform is built across three research institutions that maintain 326, 215, and 176 subjects, respectively and each institution has ten computation nodes. We perform the DDPP\_GL+DBCD along a sequence of 100 parameter values equally spaced on the linear scale of $\lambda/\lambda_{\max}$  from 1.00 to 0.1. As a comparison, we run the state-of-the-art distributed solver ADMM \cite{boyd2011distributed} with the same experiment setup. The group size is set to be 20 and we vary the number of features by randomly selecting 0.5 million to 5.9 million from GWAS dataset and report the result in Fig \ref{fig:2}.  The proposed method achieved a 38-fold speedup compared to ADMM.

\subsection{Stability selection for top risk genetic factors}
We employ stability selection \cite{li2016large,meinshausen2010stability} with DDPP\_GL+DBCD to select top risk SNPs from the entire GWAS data set with 5,906,152 features. We conduct two different groups of trials by choosing the volume of hippocampus and entorhinal cortex at baseline as the response variable for each group, respectively. In each trial, DDPP\_GL+DBCD is carried out along a 100 linear-scale sequence of parameter values from 1 to 0.05, respectively. Then we select the top 10000 features and perform stability selection \cite{meinshausen2010stability} to rank the top risk SNPs for AD. As a comparison, we perform D\_EDPP+F\_LQM  \cite{li2016large} with the same environment setup and report the result in Table \ref{tab:1}. In both of trials, \textit{APOE} is ranked 1st while DDPP\_GL+DBCD detects more risk genes like \textit{GRM8}, \textit{GPC6}, \textit{PIK3C2G} and \textit{LOC100506272} that are associated with the disease AD in GWAS \cite{rouillard2016harmonizome}. 

\begin{table}[t]
\centering
\caption{Top 5 selected SNPs with the volume of entorhinal cortex and hippocampal. }
\label{tab:1}
\centering
\begin{tabular}{|C{0.7cm}|C{0.7cm}|C{1.7cm}|C{2.3cm}|C{0.7cm}|C{0.7cm}|C{1.7cm}|C{2.3cm}|} \hline
       \multicolumn{4}{|c|}{Hippocampus by D\_EDPP+F\_LQM}& \multicolumn{4}{c|}{Hippocampus by DDPP\_GL+DBCD } \\ \hline
  No. & Chr  & RS\_ID  & Gene &No.  & Chr  & RS\_ID & Gene\\ \hline
  1 & 19  & rs429358 & APOE & 1 & 19  & rs429358 & APOE\\ \hline
  2 &  8  & rs34173062 & SHARPIN & 2 & 7  & rs1592376 & GRM8 \\ \hline
  3 &  6   & rs71573413 & unknown & 3 & 5  & rs6892867 &   LOC105377696 \\ \hline
  4 & 11  & rs10831576 & GALNT18 & 4 & 6 & rs71573413 & unknown  \\ \hline
  5 & 9  & rs3010760 & unknown & 5 & 13 & rs7317246 & GPC6 \\ \hline
\end{tabular}
\centering
\begin{tabular}{|C{0.7cm}|C{0.7cm}|C{1.7cm}|C{2.3cm}|C{0.7cm}|C{0.7cm}|C{1.7cm}|C{2.3cm}|} \hline
       \multicolumn{4}{|c|}{Entorhinal by D\_EDPP+F\_LQM}& \multicolumn{4}{c|}{Entorhinal by DDPP\_GL+DBCD} \\ \hline
  No. & Chr  & RS\_ID  & Gene &No.  & Chr  & RS\_ID & Gene\\ \hline
  1 & 19  & rs429358 & APOE & 1 & 19  & rs429358 & APOE\\ \hline
  2 & 15  & rs8025377 & ABHD2 & 2 & 4  & rs1876071 & LOC100506272\\ \hline
  3 & Y & rs79584829 & unknown & 3 & 18  & rs4486982 & unknown \\ \hline
  4 & 14  & rs41354245 & MDGA2 &  4 & 14  & rs41354245 & MDGA2  \\ \hline
  5 & 3 & rs55904134 & unknown & 5 & 12  & rs12581078 & PIK3C2G \\ \hline
\end{tabular}
\end{table}




\bibliographystyle{splncs03}

\bibliography{miccai}
\end{document}